\documentclass[]{article}
\usepackage{times}
\usepackage{cite}
\usepackage{amsmath}
\usepackage{graphicx}
\usepackage{latexsym}
\usepackage{subfigure}
\usepackage{pstricks}
\usepackage{setspace}
\begin{document}
\title{A DIKW Paradigm to Cognitive Engineering}
\author{\textbf{Amit Kumar Mishra}\\
Electrical Engineering Department \\
University of Cape Town, South Africa\\
Email: akmishra@ieee.org
}

\maketitle
\begin{abstract}
Though the word ``cognitive'' has a wide range of meanings we define cognitive engineering as learning from brain to bolster engineering solutions. 
However, giving an achievable framework to the process towards this has been a difficult task. 
In this work we take the classic data-information-knowledge-wisdom (DIKW) framework to set some achievable goals and sub-goals towards cognitive engineering. 
A layered framework like DIKW aligns nicely with the layered structure of pre-frontal cortex. 
And breaking the task into sub-tasks based on the layers also makes it easier to start developmental endeavours towards achieving the final goal of a brain-inspired system. 
\emph{\textbf{machine learning, cognitive architecture, bio-inspired, Big-Data}}
\end{abstract}
\maketitle
\section{Introduction}
Brain has intrigued researchers since the beginning of scientific endeavours. 
Firstly, beginning of computers saw the advent of exciting developments which culminated to the development of the new discipline of artificial neural networks (ANN). 
ANNs have been through several generations of major developments, with the recent phase consisting of spiking neural networks based works \cite{brans_99_book}. 
Another parallel field of computational neuroscience has been the bio-inspired cognitive architectures (BICA) \cite{sam_10_bica} a field which got major thrust in development. 
Cognitive architecture (CA) in general and BICA in particular also has a long history and the efforts have been devoted towards trying to emulate the functioning of brain. 
CAs like SOAR and ACT-R have been under development for many decades and have been applied in various studies \cite{rose_91_soar, taat_06_actr}. 
A third direction in cognitive engineering has been the recent developments in communication and radar which are misleadingly termed cognitive communication \cite{mitola_00} and cognitive radar \cite{hay_06_cograd}. 
It may also be mentioned here that this 2012-2013 has seen multi-billion dollar investment done separately in the European Union as well as in the USA for the study and understanding of brain \cite{bluebrain, human_brain, kno_13_brain}. 

With the starting of few major big-science projects \cite{human_brain,bluebrain, kno_13_brain} to understand human brain, the focus on cognitive architecture has enhanced. 
The other innovation that has made phrases like cognitive and AI the buzz-words once more is the success obtained by deep learning \cite{lecun_15_deep}. 
Another reason creating an intense interest in brain inspired architecture and hardware is the emergence of big data and the hope that this might be solved using cognitive architectures \cite{hur_15_cog_big}. 
The robotics community is also excited about the possibilities of cognitive robots \cite{rob_vision}. 
Major research agencies and industry labs are investing substantial amount of funds to explore this area. 

In spite of many very well written documents the author feels the lack of a single holistic model for cognitive engineering which is:
\begin{itemize}
\item intuitive to understand;
\item shows action-items on how to implement; 
\item shows a long term vision and how that is connected to what can be achieved now; and
\item not too abstract while not being restrictive (and thereby leaving scopes for disruptive innovations and discoveries). 
\end{itemize}

In this work we endeavour to present a model that matches the requirements listed above. 
In doing so we chose the data-information-knowledge-wishdom (DIKW) model of representing human cognition. And build our proposed cognitive architecture layer by layer. We also present the action blocks that shows how each of the layers can actually be implemented. 

Rest of the report is presented as follows. Section 2 presents our model. Section 3 expands the model with action/execution blocks. We end the report with the Conclusion section.

\section{The DIKW Model of Cognition}
From the seminal works by Fuster \cite{fuster_88} to the success of the current generation of deep neural networks \cite{pinto_11_mit} evidences suggest that the cognitive abilities of human brain emerge out of a layered architecture of the prefrontal cortex. 
Having arbitrary layers will make it untraceable and confusing. Hence, we propose to have a layered architecture inspired by the data-information-knowledge-wishdom (DIKW) framework. 
The DIKW has its own set of merits and demerits as a model. 
For our purpose of cognitive engineering this appears to be the best available in the open literature. 

\begin{figure*}[t]
	\begin{center}
		\includegraphics[width=4.5in]{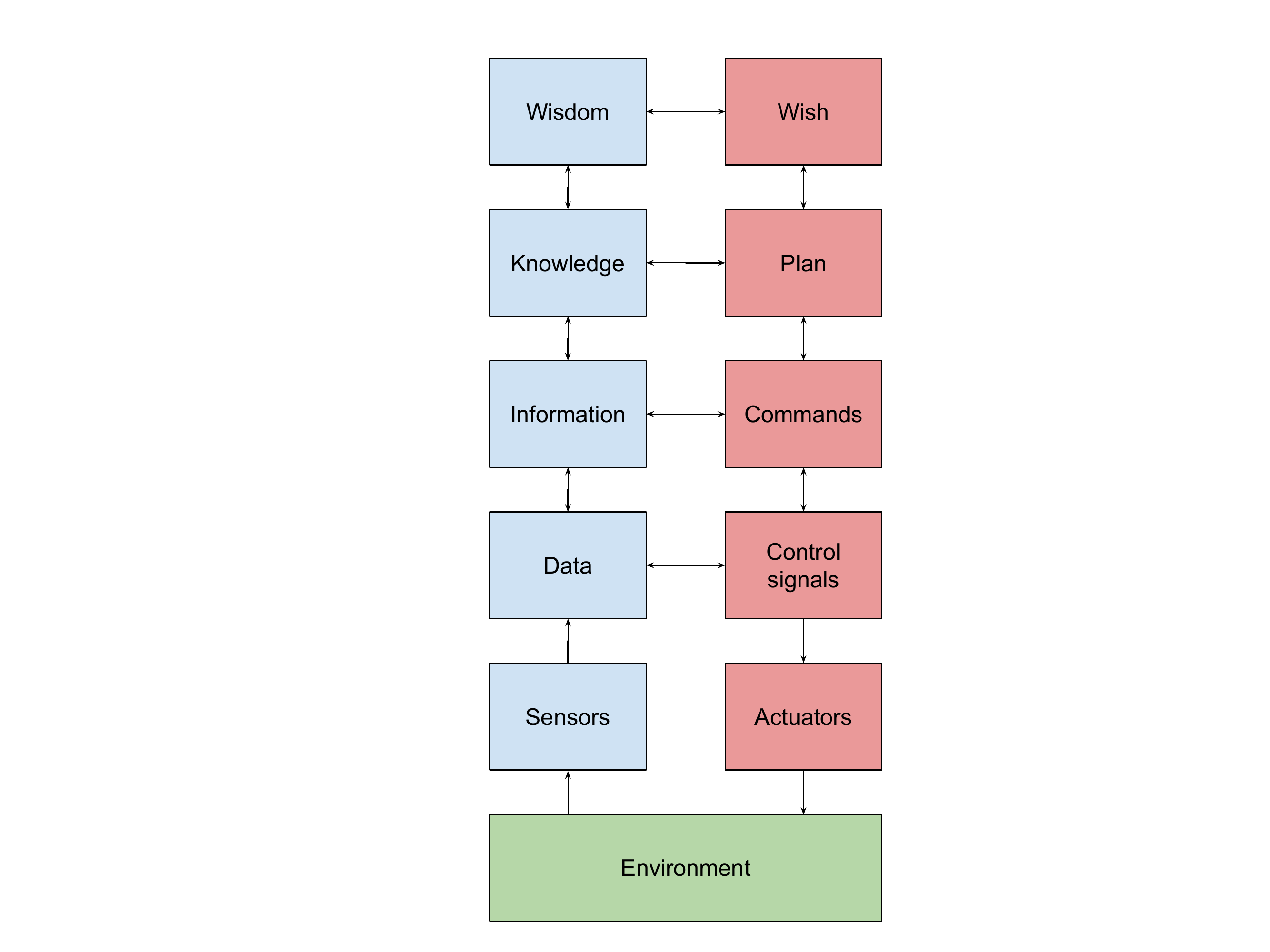}
		\caption{The DIKW model of cognitive processing. Note that each layer can be regarded as a layer in the prefrontal cortex. A layer above another represents more abstract way of processing.}
		\label{dikw_1}
	\end{center}
\end{figure*}

Figure ~\ref{dikw_1} shows the block level representation of the proposed model. 
There is a perception  path (the left hand column) that shows the path in which signal from environment passes on to the higher levels. 
The action path (the right hand column) shows the way action signal passes on from the heights level to actuator level. 
Before explaining each of the blocks we shall put forward the two major novelties in our approach. 
\begin{enumerate}
\item We have an action block associated with each of the signal processing blocks showing different levels of abstraction. 
\item We follow robotics phrases to describe many of the blocks, viz. ``plan'', ``commands'' etc. This is to give a physical interpretation of each block. It can be noted that this is by no means meant to limit the model for cognitive robotics domain only. With expanded meanings of the phrases the architecture can be applied to any generic cognitive engineering task. 
\end{enumerate}

We shall expound the different blocks of the architecture now. In doing show we shall approach it from top to bottom, i.e. from more abstract levels to less abstract ones. 
\begin{itemize}
\item {\bf Wisdom Level:} Compared to the rest three Wisdom, in DIKW, is the one with hardly any proper definition. 
To take care of this lacuna we deal with this layer with the abstract phrase ``wish'' or ``desire''. 
So the information that helps us to generate desire can be termed as belonging to this layer. 
In the data column this can be modelled by different emotions and from which may arise a ``wish'' to do something. 
This wish initiates the complete action column. 

\item {\bf Knowledge Level:} Knowledge is also very vaguely defined in the existing literature. 
We model knowledge as the ability to link disjoint bits of informations and labels. 
It can be pointed out here that this problem of linking disjoint bits of information is the focus of many recent efforts \cite{dan_17}. \\
The action column equivalent of knowledge (if we use the explanation of knowledge we discussed just now) is ``planning''. 
Planning needs a holistic view of the situation and the task and hence needs link between all disjoint bits of information. 
It can also be noted that planning as a task is well accepted as that needing higher cognitive abilities \cite{hal_10_plan_cog}. 

\item {\bf Information Level:} 
In spite of many well accepted definitions of this word the exact meaning of information changes from domain to domain. 
The Oxford Dictionary defines information as ``facts provided about something; what is conveyed or represented by a particular arrangement or sequence of things''. \\
From this generic definition (and we should not forget that we are developing a generic architecture which should be modifiable to different domains) we can use ``information'' to mean labels attached to blocks of data conveying certain meanings. 
For example, a block of pixels can be converted to a label called ``flower''. \\
The action column equivalent of information is command. 
For example in robotics it can be ``turn left'' or ``slow down by 20\%''. 

\item {\bf Data Level:} This is the first level and consists of data as collected from the environment through sensors. \\
In the action column this will consist of actuators. 

\end{itemize}

\section{Achieving Cognition through the DIKW Model}

The model explained in the above section is a complicated one. 
In this section we shall expand the block diagram by including processing blocks which will convert the output from one block to the input of another. 

A limitation of the current work is the fact that some action blocks are abstract which means we can not implement them using the current stage of technologies. 
However the beauty of a modular architecture like the current one is that we can leave the upper blocks and implement whatever levels we can starting from the lowest level. 

\begin{figure*}[t]
	\begin{center}
		\includegraphics[width=4.5in]{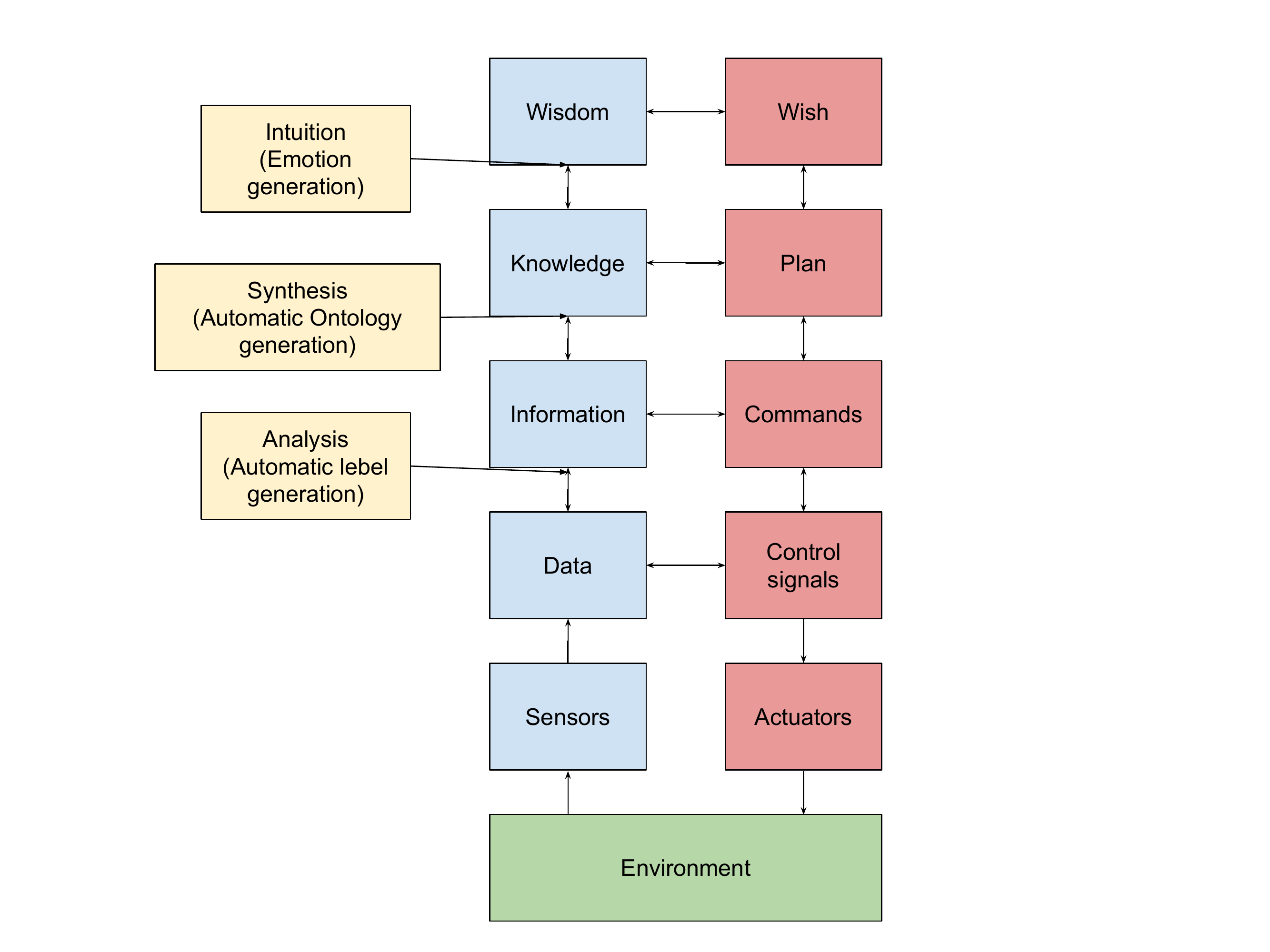}
		\caption{Just a model is not of much use unless there is plan for implementing the same. This figure shows additional action-blocks which can, potentially, transform D to I, I to K and K to W. It can however be noted that the last transformation of K to W is a proposal for completion sake. The authors are not aware of any work endeavouring to extract W from K; even the definition of wisdom is far from set in stone.}
		\label{dikw_2}
	\end{center}
\end{figure*}

Figure ~\ref{dikw_2} gives the block diagram of the implementation oriented architecture with action blocks shown in yellow. 

We shall explain each action block below with reference to their strata in the DIKW chain. 
\begin{itemize}
\item {\bf D-I Interface:} We term this action block as ``analysis''. 
This is one of the exhaustively researched area in the recent past. 
As we have discussed in the previous section, extracting information can be understood as attaching labels to the data. 
This is the set of algorithms used for pattern classification. Hence, this block is one of the easier-to-implement blocks. 

\item {\bf I-K Interface:} As discussed in the previous section, knowledge can be taken as the steps in which related (and possibly disjoint) labels are linked together. 
We call the execution steps that can give this functionality as ``synthesis''. 
This is a field of current research. 
{\it Automatic ontology extraction} \cite{} is a field of research whose algorithms can be borrowed to execute this block. 
However, ontology itself is a fairly recent field of research and automatic ontology is still in its infancy. \\
Still the visions are set clear and we expect the execution of this block to be feasible in the coming few years. 

\item {\bf K-W Interface:} This is the most abstract execution block. 
We call it the ``Intuition'' block which generates wisdom from knowledge (or ontology). 
We foresee that this block will mostly involve the generation of emotions. 
However, the exact meaning of emotions and the exact modus-operandi of their generation is still something that may take a decade to develop. So we can hope that this block can only be implemented in the coming decade. 
\end{itemize}

\begin{figure*}[t]
	\begin{center}
		\includegraphics[width=4.5in]{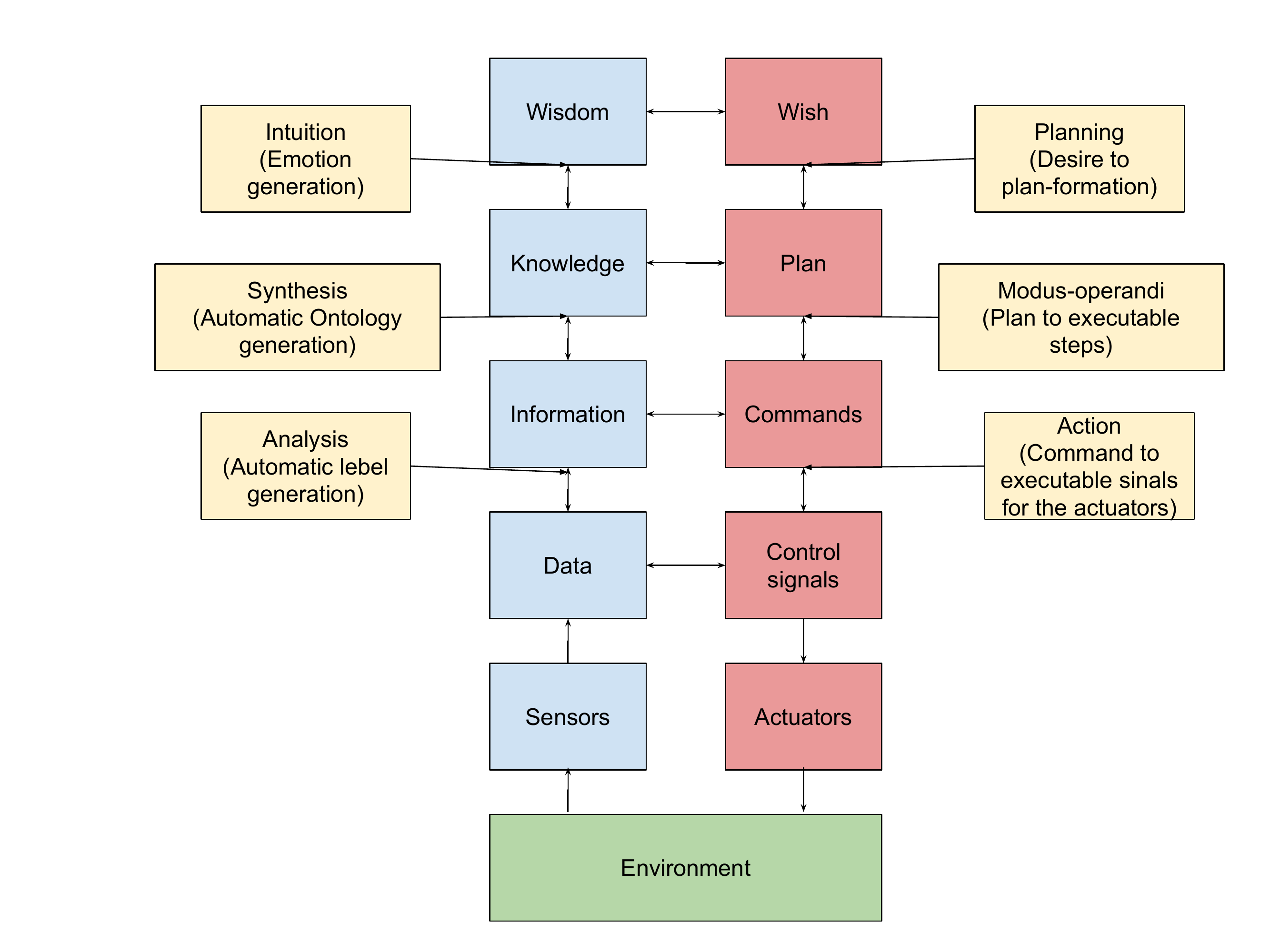}
		\caption{The architecture with executable blocks for the action column as well.}
		\label{dikw_3}
	\end{center}
\end{figure*}

Next we add another layer to the architecture where we put execution blocks for the action column as well. 
This is shown in Figure~\ref{dikw_3}. 
We shall explain each action block below with reference to their strata in the DIKW chain. 
\begin{itemize}
\item {\bf W-K Interface:} We call this block planning. 
This takes desire to the level of plan-formation stage. 
Again, being at the top-most layer, the execution of this block will depend on the detailed understanding of ``desire'' and ``wisdom'' and hence may take almost a decade. 

\item {\bf K-I Interface:} We call this block ``Modus-operandi'' in which the plan is broken into executable steps. 
This is something that can be executed with the current technology. 
Of course it depends on the domain. For example, in the domain of path-planning this is a well-studied area (though, by no means, exhausted). However, in a different domain (e.g. financial asset management) this will still take a while to be executable. 

\item {\bf I-D Interface:} This is the last execution block where the commands are broken into executable signal that can be understood by the actuators. 
This, again, is an implementable block with the current technology. 

\end{itemize}

\section{Conclusion}
In this work, we have presented the model of prefrontal cortex from a DIKW paradigm. 
Prefrontal cortex is attributed to cognition in human brain. Hence, we believe that the implementation of this will create a truly cognitive system. 

The merit of the presented architecture is that it is based on DIKW paradigm which is well understood and hence understanding the architecture is easy. 
Secondly this approach helped us to break the architecture into distinct blocks. This helps in further analysis of the architecture and helps in implementing this. 

Lastly, we also preset execution blocks which will make the complete architecture possible to be implemented.  
It can be marked that almost three of the execution blocks can not be implemented using the current technology. 
However the beauty of the architecture is the fact that because of the layered nature we can still implement part of it and experiment. This also helps in setting a road-map for future development which will help in implementing the architecture fully. 

We believe that the proposed architecture will help the researchers in implementing partially cognitive systems and will also set a trackable road-map for the cognitive engineers.

\bibliographystyle{IEEEtran}
\bibliography{ref}

\end{document}